**Title Page**

**Original Article**

**Employing General-Purpose and Biomedical Large Language Models with Advanced Prompt Engineering for Pharmacoepidemiologic Study Design**

Xinyao Zhang[1*] & Nicole Sonne Heckmann[1*], Manuela Del Castillo Suero[1], Francesco Paolo Speca[1], Maurizio Sessa[1]

[1]Department of Drug Design and Pharmacology, University of Copenhagen, Copenhagen, Denmark.

*Equally contributed as first-author.

**Metrics**

*Manuscript*: 3076

*Abstract*: 178

*References*: 27

*Figures and tables*: 1 table and 6 figures

**Corresponding author:**

Maurizio Sessa,

Department of Drug Design and Pharmacology,

University of Copenhagen,

Jagtvej 160, 2100 Copenhagen, Denmark

Email: maurizio.sessa@sund.ku.dk

**Abstract**

**Background:** The potential of large language models (LLMs) to automate and support pharmacoepidemiologic study design is an emerging area of interest, yet their reliability remains insufficiently characterized. General-purpose LLMs often display inaccuracies, while the comparative performance of specialized biomedical LLMs in this domain remains unknown.

**Methods:** This study evaluated general-purpose LLMs (GPT-4o and DeepSeek-R1) versus biomedically fine-tuned LLMs (QuantFactory/Bio-Medical-Llama-3-8B-GGUF and Irathernotsay/qwen2-1.5B-medical_qa-Finetune) using 46 protocols (2018-2024) from the HMA-EMA Catalogue and Sentinel System. Performance was assessed across relevance, logic of justification, and ontology-code agreement across multiple coding systems using Least-to-Most (LTM) and Active Prompting strategies.

**Results:** GPT-4o and DeepSeek-R1 paired with LTM prompting achieved the highest relevance and logic of justification scores, with GPT-4o-LTM reaching a median relevance score of 4 in 8 of 9 questions for HMA-EMA protocols. Biomedical LLMs showed lower relevance overall and frequently generated insufficient justification. All LLMs demonstrated limited proficiency in ontology-code mapping, although LTM provided the most consistent improvements in reasoning stability.

**Conclusion:** Off-the-shelf general-purpose LLMs currently offer superior support for pharmacoepidemiologic design compared to biomedical LLMs. Prompt strategy strongly influenced LLM performance.

# 1. Introduction

The growing need for rapid evidence generation in pharmacoepidemiology places strong demands on study design and data quality.[1] This is particularly important for populations underrepresented in clinical trials and comparative effectiveness/safety research in routine care. [2,3] As artificial intelligence (AI) becomes increasingly integrated into biomedical research, large language models (LLMs) have been proposed as tools to support time- and resource-consuming tasks, including those relevant to pharmacoepidemiology. [4–6] Interest in this area is growing within regulatory settings. The U.S. Food and Drug Administration (FDA) has explored the use of LLMs such as BERTox and GPT to support document analysis, literature review, and toxicity assessment. [5] The European Medicines Agency (EMA) has issued high-level recommendations on the use of LLMs in regulatory activities. [6] Despite this momentum, the role of LLMs in time- and consuming-tasks in pharmacoepidemiology, such as study design, remains unclear. Pharmacoepidemiologic study design requires accurate identification and formulation of multiple interrelated methodological elements, including temporal assessment windows, inclusion and exclusion criteria, exposure definition, and ontology-code mapping. [7–9] Early attempts to use off-the-shelf LLMs for these tasks have highlighted limitations. [10–12] However, it remains uncertain whether such limitations are mainly related to the type of model (i.e., general-purpose), prompt strategy, or both. This study therefore aimed to evaluate the performance of off-the-shelf general-purpose LLMs and biomedical LLMs, combined with prompt-engineering strategies, for supporting pharmacoepidemiologic study design.

# 2. Methods

## 2.1 Data Sources

This study evaluated LLM performance using study protocols from two pharmacoepidemiologic sources (i.e., HMA-EMA Catalogue and Sentinel System). From the HMA-EMA Catalogue of Real-World Data (RWD) Studies, all available DARWIN EU® protocols were included at the time of analysis (n=16). In addition, 15 non-DARWIN expert-developed protocols were randomly sampled from the HMA-EMA Catalogue and were required to be registered between January $1^{st}$, 2023, and February 22nd, 2024, finalized, written in English, and publicly available. From the Sentinel System, 15 protocols were selected registered between January 1st, 2022, and February 22nd, 2024, marked as complete, classified as "drug and outcome analysis" or "drug use," and publicly available. [13,14]

### 2.2 Pharmacoepidemiologic Design Concepts

The evaluation framework was structured around key pharmacoepidemiologic design concepts derived from the HARPER checklist. [15] These concepts were operationalized as a sequence of concept-specific tasks to reflect the main elements typically required for protocol design in RWD studies. The concept-specific tasks were structured as follows: a) Question 1 assessed the correct identification of the study design; b) Question 2 assessed the definition of the index date; c) Question 3 assessed inclusion and exclusion criteria; d) Question 4 assessed the inclusion and exclusion assessment window; e) Question 5 assessed exposure; f) Question 6 assessed the outcome; g) Question 7 assessed the follow-up period; h) Question 8 assessed covariates; i) Question 9 assessed the covariate assessment window; and then it was separately assessed ontology-code mapping.

### 2.3 Human Expert Gold Standard

From each protocol, 2 human assessors (XZ and MS) extracted pharmacoepidemiologic design concept mentioned in section 2.2, which served as the gold standard against which LLM outputs were evaluated. Specifically, they included study design, index date, inclusion and exclusion criteria, inclusion and exclusion assessment window, exposure, outcome, follow-up period, covariates, covariate assessment window, and ontology-code mapping. [15] LLM-generated outputs were compared with protocol-derived expert answers across the 10 questions (Table 1 provides the specific prompts used for each question).

### 2.4 Prioritization of Large Language Models and Prompt Strategies

Two general-purpose LLMs (GPT-4o and DeepSeek-R1) and four biomedical LLMs were included in the screening phase: QuantFactory/Bio-Medical-Llama-3-8B-GGUF, mradermacher/DeepSeek-r1-Medical-Mini-GGUF, Plaban81/gemma-medical_qa-Finetune, and Irathernotsay/qwen2-1.5B-medical_qa-Finetune. [16–19] Model selection for biomedical LLM was based on overall performance as reported by the Open LLM Leaderboard hosted on Hugging Face (https://huggingface.co/spaces/open-llm-leaderboard/open_llm_leaderboard#/) as of March 2025. LLMs were paired with seven prompt engineering strategies identified from the literature: 1) basic prompt with term definitions, 2) synthetic prompting, 3) active-prompt, 4) plan-and-solve, 5) least-to-most (LTM), 6) Tree-of-Thought, and 7) decomposition prompting. [20–22] Together, this yielded 42 model-prompt combinations for the initial pre-assessment of the most performing combinations. Prompt construction was guided by the CLEAR framework, which emphasizes concise, logical, explicit, adaptive, and reflective prompt design. [23] A pre-assessment was conducted on three randomly selected protocols, one from each protocol source. Model-prompt combinations were evaluated against the human expert gold standard using a 1-5 Likert scale for relevance and logic of justification, a binary agreement measure for logic of justification, and error frequency as an additional criterion. Based on this screening, the five top-performing combinations were selected for the main analysis: GPT-4o with LTM, GPT-4o with Active Prompt, DeepSeek-R1 with LTM, QuantFactory/Bio-Medical-Llama-3-8B-GGUF with LTM, and Irathernotsay/qwen2-1.5B-medical_qa-Finetune with LTM. Protocols, extracted data, and LLM-generated output from the pre-assessment are available at: https://github.com/madelsu/LLM-for-pharmacoepi-study-design/tree/main/Data/pre_assessment.

## 2.5 Study Outcomes

The primary outcome was relevance, defined as agreement between LLM-generated answers and protocol-derived expert answers measured with Likert-scale. Agreement was considered acceptable if scores were identical or differed by no more than one Likert level. Logic of justification was assessed as a key secondary outcome using a 5-point scale, with 1 indicating illogical or inconsistent reasoning and 5 indicating coherent, well-supported, and methodologically rigorous reasoning. Ontology-code agreement was assessed by comparing predicted and reference codes (i.e., from the protocols) for exposure and outcome across ontology groups and summarizing percentage agreement by model, prompt strategy, and protocol source. Error patterns were examined through qualitative review of predefined error types, and frequencies were summarized descriptively.

## 2.6 Data Analysis and Software

A descriptive analysis of the included protocols was performed by protocol source (DARWIN EU®, HMA-EMA, and Sentinel System). Counts and proportions were calculated for publication year, geographical origin of authors, study type, database use, and ontology-code mappings used to define exposure and outcome. For relevance and logic of justification, Likert-scale results were summarized for each model-prompt combination using medians, interquartile ranges (IQRs), means, whisker ranges, and outlier counts. These distributions were visualized using box plots. Comparisons were made across model-prompt combinations, by LLM architecture, by prompt strategy, and by protocol source. In the box plots, white dots represented mean values, black dots indicated statistical outliers, and whiskers extended to the range within 1.5×IQR. Ontology-code agreement was calculated as percentage agreement between model-predicted and reference codes and was summarized by ontology-code group, model, prompt strategy, and protocol source. Results were visualized using comparative bar plots. Error frequencies were tabulated and summarized descriptively. To explore error patterns across questions and biomedical LLMs, deviations from the mean occurrence of each error type per question were calculated and visualized using a heatmap. All statistical analyses and visualizations were performed in R version 4.5.1. Data handling and plotting were conducted using tidyverse version 2.0.0, box plots and bar charts were generated using ggplot2 version 3.5.1, and heatmaps were created using the heatmap package version 1.0.12.

## 2.7 Reporting Standards

This article was prepared in accordance with the TRIPOD-LLM and CHART reporting standards (Supplementary Tables 2 and 3). [24,25]. All statistical analysis code, study protocols, LLM prompt strategies, relevance and logic of justification assessments are publicly available via a GitHub repository (Link: https://github.com/madelsu/LLM-for-pharmacoepi-study-design.git).

## 3. Results

### 3.1 Descriptive Analysis

A total of 46 pharmacoepidemiological protocols were included: 16 from DARWIN EU®, 15 from the HMA-EMA Catalogue, and 15 from the Sentinel System. The 46 protocols are available at https://github.com/madelsu/LLM-for-pharmacoepi-study-design/tree/main/Data. There was full agreement between the human experts when extracting information from the protocols. The protocols were published between 2018 and 2024, with annual frequencies of 1 from 2018, 2 from 2019, 4 from 2020, 3 from 2021, 3 from 2022, 29 from 2023, and 4 from 2024. DARWIN EU® and HMA-EMA protocols involved researchers from Belgium, Estonia, Finland, France, Germany, Italy, the Netherlands, Norway, Romania, Spain, and the United Kingdom, whereas Sentinel protocols originated from the United States. Study topics were primarily drug utilization (n=18) and disease epidemiology (n=15), followed by safety (n=11) and effectiveness (n=2). Across the included protocols, 15 data sources were represented, of which IQVIA Germany (n=23), SIDIAP (n=15), and the Sentinel Distributed Database (n=14) were the most frequently used. Fifteen ontology-code systems were identified, with ICD, SNOMED, ATC, CPT, and RxNorm among the most frequently used for concept definition. Less frequently used systems included MedDRA and Read codes.

### 3.2 Relevance

Overall, GPT-4o-LTM and DeepSeek-R1-LTM achieved the highest relevance scores. GPT-4o-LTM showed the most stable performance, reaching a median score of 4 with a narrow IQR in 8 of 9 questions for HMA-EMA protocols. Across evaluated concepts, general-purpose LLMs outperformed the biomedical LLMs, with higher median relevance scores and greater overall stability (Figure 1). LTM prompting also generally performed better than Active Prompting, particularly for study design and exposure-related tasks, where it produced higher scores and narrower score distributions (Figure 2). Protocol sources influenced model performance. HMA-EMA protocols yielded the highest and most stable relevance scores, particularly for GPT-4o-LTM and DeepSeek-R1-LTM. Sentinel protocols also supported relatively strong performance but with greater variability. In contrast, DARWIN EU® protocols were associated with lower and less stable relevance, especially for outcome and follow-up definitions. Across all protocol sources, biomedical LLMs showed lower relevance overall, with DARWIN EU® protocols yielding consistently low-scoring outputs (Figure 3).

### 3.3 Logic of Justification

The overall pattern for logic of justification largely mirrored the relevance results. GPT-4o-LTM and DeepSeek-R1-LTM again achieved the strongest performance, with GPT-4o-LTM on HMA-EMA protocols reaching the highest and most stable scores across questions. Non-biomedical LLMs generally produced more coherent and methodologically plausible justifications than biomedical LLMs, which were more frequently not assessable because of missing or inadequate reasoning (Figure 4). As in the relevance analysis, HMA-EMA protocols yielded the most stable performance, whereas Sentinel and DARWIN EU® protocols showed greater variability across concepts (Figure 5). LTM prompting generally outperformed Active Prompting, particularly for study design, exposure, and outcome definition tasks. However, LTM also produced occasional low-end outliers, suggesting that errors introduced in early steps could propagate across later dependent responses (Figure 6).

### 3.4 Ontology-Code Agreement

Ontology-code agreement was markedly lower than performance on the narrative design tasks. Across protocol sources, agreement was observed mainly for ATC and selected procedure-based coding systems, whereas performance was limited for ICD-based systems and absent or negligible for RxNorm and SNOMED in several settings. The highest agreement was observed for ATC coding in HMA-EMA protocols with DeepSeek-R1-LTM (28.6%) and for HCPCS coding in Sentinel protocols with GPT-4o-LTM (22.2%). Overall, ontology-code mapping remained a weak area across all LLMs.

### 3.5 Error Classification

Error analysis showed that the biomedical LLMs were particularly prone to recurrent structured failures. The most frequent error types were irrelevant auto-generated questions (n=189) and lack of justification (n=105), followed by long answers (n=36) and non-adherence to the intended algorithm (n=11). Less frequent errors included prompt echoing, model crash, and language inconsistency. Overall, QuantFactory/Bio-Medical-Llama-3-8B showed more concentrated error clusters across core questions, particularly Questions 3-7, whereas Qwen displayed a flatter but still suboptimal error profile (Supplementary Figure 1). See Supplementary Table 1 for error definitions and examples.

## 4. Discussion

The use of LLMs in pharmacoepidemiology has received increasing research attention in recent years.[11] Despite this growing interest, important concerns remain regarding the reliability of general-purpose LLMs, the suitability of their training data for scientific and regulatory tasks, and the performance of newly released models and prompt-engineering strategies in supporting pharmacoepidemiologic study design. These uncertainties highlight the need for systematic evaluation of LLM capabilities in this domain. This study addressed that need by evaluating off-the-shelf general-purpose LLMs, biomedical LLMs, and advanced prompt-engineering strategies for support of pharmacoepidemiologic study design. A total of 46 study protocols from DARWIN EU®, the Sentinel System, and HMA-EMA were assessed by both human experts and LLMs. Performance was assessed across multiple dimensions, including relevance, logic of justification, ontology-code agreement, and error patterns.

The main finding was that general-purpose LLMs, particularly GPT-4o and DeepSeek-R1 paired with LTM prompting, consistently outperformed the biomedical LLMs evaluated in this study.

This pattern was observed not only for the primary outcome of relevance, but also for logic of justification, suggesting that the advantage of these LLMs extended beyond answer selection to broader methodological reasoning. Although the biomedical LLMs were domain-specific, they showed lower relevance, weaker justification quality, and more frequent structured failures, particularly irrelevant auto-generated questions and lack of justification. This suggests that domain adaptation alone may be insufficient for tasks that require not only biomedical knowledge, but also structured interpretation of study design elements such as index date, follow-up, covariates, and outcome definitions. In other words, the task appears to depend as much on methodological reasoning and contextual interpretation as on domain vocabulary. This is consistent with findings reported by Heckmann et al. in pharmacovigilance, where general-purpose LLMs also outperformed biomedical LLMs on tasks requiring structured reasoning and explainable output. [26] It is also in line with Dada et al., who showed that increasing biomedical specialization does not necessarily translate into better performance on medical tasks and may come at the cost of instruction-following or broader task adaptability. [27] Together, these findings suggest that pharmacoepidemiologic study design is not simply a knowledge retrieval task, but one that requires LLMs to understand instructions and organize interrelated pharmacoepidemiological design choices in a coherent way. The stronger performance of general-purpose LLMs therefore suggests that, for this type of task, broader semantic understanding may currently be more important than narrow biomedical tuning.

Prompt strategy also influenced LLM performance. LTM prompting generally produced more relevant and better-justified outputs than Active Prompting, particularly for study design, exposure, and outcome-related tasks. This likely reflects the advantage of decomposing complex pharmacoepidemiologic reasoning into sequential subtasks, allowing LLMs to build more complete and coherent reasoning chains. For tasks that involve multiple dependent design elements, such structured prompting may be particularly useful. At the same time, the persistence of low-end outliers indicates that this structure may also increase sensitivity to early-step errors, which can then propagate into later responses. Prompting therefore improved performance but did not remove the need for caution or review. This is especially relevant in workflows where later answers depend directly on earlier ones, as was the case for several of the pharmacoepidemiologic concepts assessed in this study.

A notable finding was the influence of protocol source and format. HMA-EMA protocols yielded the highest and most stable scores, while Sentinel protocols showed broader variability, and DARWIN EU® protocols tended to produce lower and less stable performance. Because this pattern was seen across model types, protocol structure itself appears to shape model performance. This suggests that LLM outputs are influenced not only by model architecture and prompting strategy, but also by the clarity, consistency, and reporting style of the source material. If protocol format affects model performance, then standardization is not only a reporting issue but also a deployment issue. More consistently structured protocols may reduce ambiguity for both human and machine readers and may be especially important in highly regulated settings, including EMA-related contexts, where documentation quality, traceability, and methodological consistency are central requirements.

Ontology coding remained a clear weakness across all LLMs and protocol sources. Even when the LLMs performed well on narrative design tasks, ontology-code agreement was generally limited, particularly for more complex coding systems such as ICD, RxNorm, and SNOMED. This suggests that coding tasks place different demands on LLMs than narrative interpretation tasks and that good performance in one area does not necessarily translate into reliable performance in the other. This distinction is important because ontology coding is not a minor technical detail in pharmacoepidemiologic work. It is part of how exposure, outcome, and covariate definitions become operationalizable and reproducible. Poor coding performance therefore has direct implications for downstream validity. This means that although LLMs may assist with interpretation and drafting, ontology coding should not rely on free-text inference alone. Deterministic support, such as curated dictionaries, crosswalks, or rule-based validation, remains necessary.

### 4.1 Strengths and Limitations

A key strength of this study is its systematic comparison across multiple dimensions, including model architecture, prompt strategy, protocol source, and ontology coding. This multidimensional framework provides a more comprehensive picture of LLM behavior in pharmacoepidemiologic contexts than would be possible from isolated performance metrics alone. Another strength is the use of human expert ratings as a benchmark, which grounds the findings in practical relevance and helps identify where automation may meaningfully support expert workflows and where risks remain. The inclusion of protocols from three major networks, DARWIN EU®, HMA-EMA, and Sentinel, further strengthens the study by allowing performance to be examined across diverse reporting standards and coding systems. In addition, the focus on both relevance and logic of justification extends the evaluation beyond surface-level accuracy and addresses the interpretability and soundness of model reasoning.

Several limitations should nevertheless be acknowledged. First, the study evaluated off-the-shelf LLMs without task-specific fine-tuning. While this reflects real-world use cases, performance may differ with domain-adapted training pipelines incorporating structured ontology coding. Second, the evaluation of logic of justification showed substantial variability, particularly for complex concepts, suggesting that the scoring framework may require further refinement to capture nuance reliably. Third, although protocols from three networks were included, they are not fully representative of all global pharmacoepidemiologic practices, and results may differ in settings with alternative reporting standards or coding systems. Finally, the reliance on Likert-scale ratings and agreement thresholds introduces subjectivity into the evaluation, even though multiple raters and comparisons were used to mitigate this limitation.

## 5. Conclusions

This study showed that off-the-shelf general-purpose LLMs, particularly GPT-4o and DeepSeek-R1 combined with LTM prompting, outperformed the biomedical LLMs evaluated in this study for support of pharmacoepidemiologic study design. Their advantage was evident not only for the primary outcome of relevance, but also for logic of justification, indicating that stronger performance was observed both in alignment with protocol-derived expert answers and in the ability to provide more coherent methodological reasoning.

## 6. Ethics, Funding, and Conflict of Interest

No funding was obtained for this study. Because the study relied exclusively on publicly available study protocols, ethical approval and informed consent were not applicable. The authors declare no conflicts of interest with respect to the conduct or reporting of this study.

## 7. Authors' Contribution

NSH and MS had primary responsibility for the study conception and design and for the acquisition, analysis, and interpretation of data, and they led the drafting of the manuscript. MS and XZ contributed to data collection, formal analysis, interpretation of results, and revision of the manuscript. MDCS and FPS contributed to interpretation of results and critical revision of the manuscript. All authors read and approved the final manuscript and agree to be accountable for the work, including ensuring that any questions related to the accuracy or integrity of the study are appropriately addressed.

**Table 1.** Prompt Examples,

| Concept | Least to Most | Active prompt |
|---------|---------------|---------------|
| Study design | Based on the title [TITLE] , list all the candidate study design types (e.g., cohort study, case-control study) and select the most appropriate one and justify it | Primary prompt:<br>As a pharmacoepidemiologist , recommend the optimal study design type for the study titled "{TITLE}" . Consider all candidate designs (e.g., cohort, RCT, case control). Constraints:<br>1. Align with CONSORT/STROBE guidelines<br>2. Prioritize feasibility in real-world settings<br>3. Explicitly state rejection reasons for alternative designs Then enhance prompts:<br>[CoT Example]<br><br>Objective: Assess long-term effects of Drug X on risk<br>Step 1: Identify need for longitudinal exposure-outcome data → Cohort design<br>Step 2: Check feasibility of randomization → Reject RCT (lack of equipoise)<br>Step 3: Compare with case-control → Prefer cohort for temporal clarity |
| Index date | Define the index date for this study based on the selected [Answer to Question 1] study design, indicating its specific definition and whether it is fixed or variable. | Primary prompt:<br>Define index dates of this study. Requirements:<br>1. For pharmacological studies: Use first prescription date + 30-day washout<br>2. For surgical studies: Use procedure date ± 7-day pre-op assessment<br>3. Flag potential immortal time bias sources<br><br>[Validation] Check FDA Sentinel Common Data Model, DARWIN, HMA-EMA for alignment |
| Inclusion and | Inclusion criteria (e.g., age ≥18 years, confirmed disease) and exclusion criteria (e.g., past | Generate inclusion/exclusion criteria of this study. Mandatory elements: |

| exclusion criteria | medical history, missing data) should be developed for this study based on the [Q1 answer] design and [Q2 answer] index date | 1. Explicit linkage to objective-specific endpoints<br>2. Use WHO International Classification of Functioning (ICF) framework<br>3. Flag criteria causing selection bias >20% |
|---|---|---|
| Inclusion and exclusion assessment window | Based on the [answer to question 2] index date, identify the time window for assessing inclusion and exclusion criteria (e.g., a baseline period of 6 months before the index date, an exclusion period of 1 year after the index date) and its specific duration | Primary prompt:<br>Define temporal assessment windows for inclusion/exclusion criteria of the answer to question 3.<br><br>Constraints:<br>1. Align with EMA Guideline on GCP (Rev 3)<br>2. Use moving window approach for chronic conditions<br>3. Specify allowable overlaps (±5% timeline tolerance) Enhance prompt:<br><br>[CoT Example]<br><br>Objective:<br>Window Logic:<br>1. Baseline:<br>2. Exclusion:<br><br>[Check] Apply inverse probability weighting for missing window data |
| Exposure | Based on the [Answer to Question 1] design and [Answer to Question 3] inclusion criteria, define exposure (e.g., drug dose, exposure duration) | Primary prompt:<br>Define study exposure of the study Required elements:<br>1. Dose-response granularity (ATC + RxNorm mapping)<br>2. Exposure lag periods with biological plausibility check<br><br>Competing risk adjustment plan |
| Outcome | Based on the [answer to question 1] design, specify primary/secondary outcome | Primary prompt:<br>Specify outcome for this study. Must include:<br>1. Primary/secondary endpoints |
| | definitions (e.g., laboratory confirmed diagnosis, imaging evidence). | 2. Competing risk definitions (e.g., death censoring rules)<br><br>Sensitivity analysis protocols for outcome misclassification. |
| Follow-up period | In conjunction with [Q.2 answer] Index date and [Q.6 answer] Definition of ending, set the start and end of the follow up period, whether right censoring is allowed and the minimum length of follow-up (e.g. 1 year). | Question 7:<br>Primary prompt:<br>Define follow-up periods in this study.<br>Requirements:<br>  1. Account for disease-specific latency periods<br>  2. Apply landmark analysis for time-varying exposures<br>  3. Specify left-truncation handling methods Enhance prompt:<br><br>[CoT Example]<br><br>Objective:<br>Follow-up Logic:<br>  1. Start:<br>  2. End:<br><br>Censor: Death |
| Covariate | List covariates (e.g., age, sex, comorbidities, medication history) to be adjusted for based on [QUESTION 5 ANSWER] exposure and [QUESTION 6 ANSWER] outcome. | Primary prompt:<br>Identify covariates in this study.<br>Must:<br>  1. Categorize into confounders/mediators/effect modifiers<br>  2. Provide LOINC codes for lab covariates<br>  3. Include negative control variables<br><br>Enhance prompt:<br>[CoT Example]<br><br>Objective:<br><br>Covariates:<br>  1. Confounder:<br>  2. Effect modifier: |
| | | Negative control: |
| Covariate assessment window | Based on the [QUESTION 2 ANSWER] index date, define the time window in which the covariates will be evaluated (e.g., laboratory data within 1 year prior to the index date). | Primary prompt:<br>Define temporal windows for covariate assessment in this study. Constraints:<br>1. Align with FDA's Structured Product Labeling (SPL) standards<br>2. Handle time-varying covariates using marginal structural models<br>3. Specify missing data imputation thresholds Enhance prompt:<br><br>[CoT Example]<br><br>Objective:<br><br>Window Strategy:<br>• Static:<br>• Time-varying:<br>Lagged: |
| Ontology-code | Convert diagnoses, exposures, outcomes from [Q5/6 answers] to standard codes i.e. ICD-10, ATC. | Map all medical concepts in previous answers to standard terminologies based on the answer to question 5, 6 list all the code of this concept.⌊SEP⌋ Required systems: |

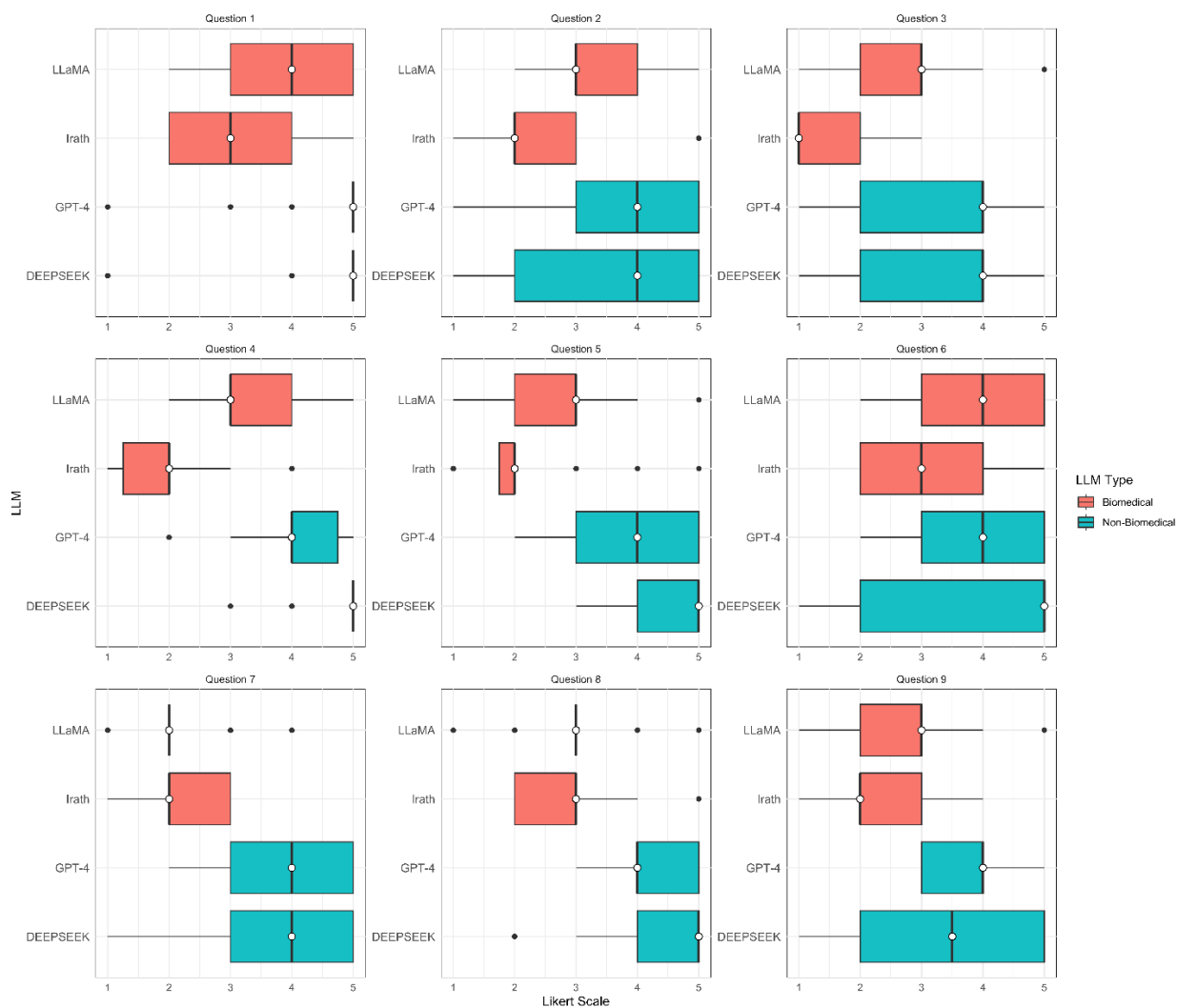

**Figure 1.** Box plots on the distribution of 5-Likert-scale relevance scores for outputs generated by different LLMs architectures across 9 questions.

*Legend: Each subplot (1-9) represents one question of specific pharmacoepidemiological study. The y-axis lists LLM–prompt combinations, while the x-axis indicates Likert scores from 1 (completely inaccurate) to 5 (highly accurate). Box plots illustrate the interquartile range (IQR), median (horizontal line within the box), and overall score dispersion. White dots represent the mean (average) accuracy score for each combination, providing a complementary measure of central tendency. Black dots denote statistical outliers, reflecting individual responses that deviate substantially from the main distribution. The whiskers extend to the minimum and maximum values within 1.5×IQR of the box (non-outlier range). Red box reflects biomedical LLMs, and blue box presents non-biomedical LLMs.*

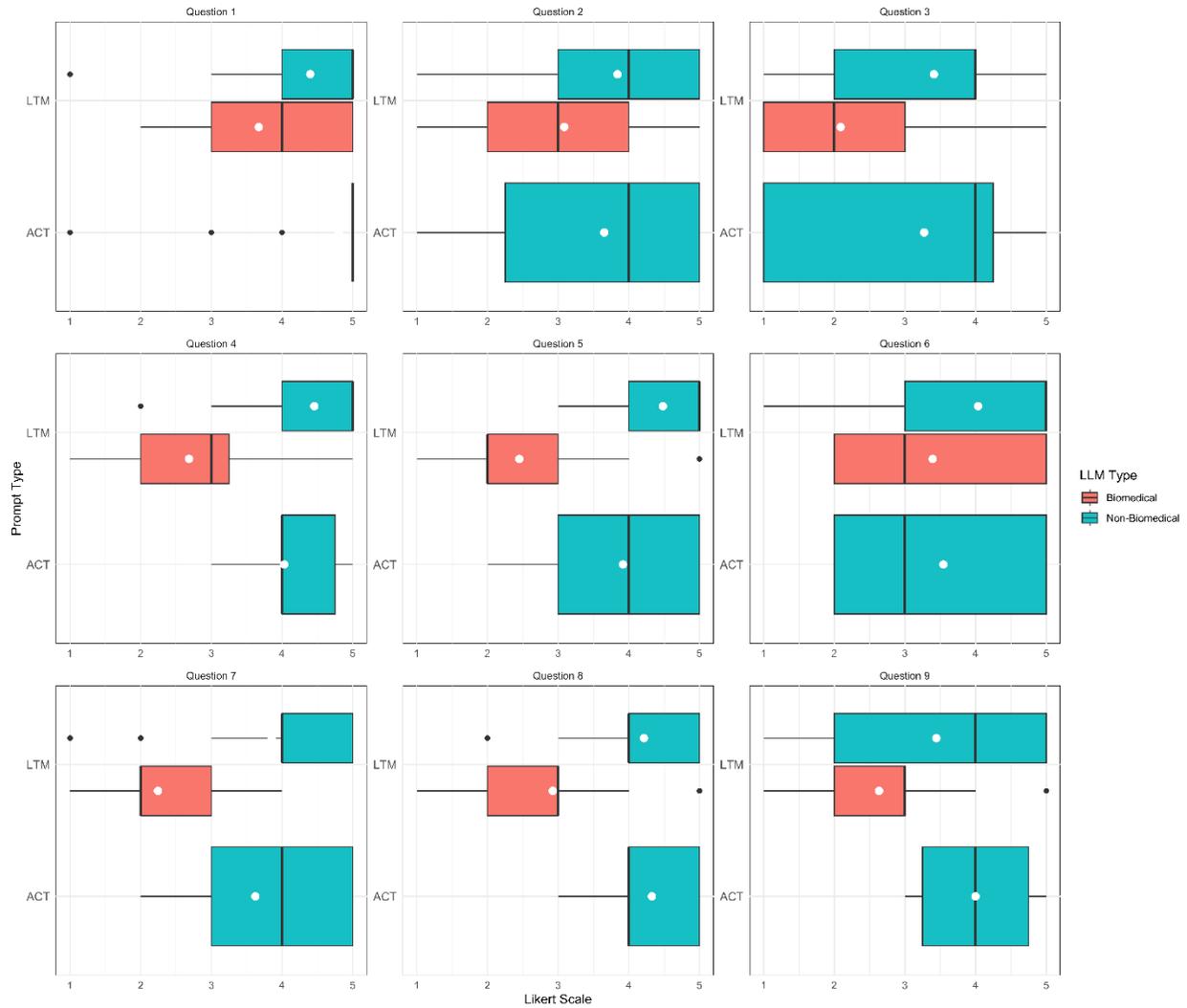

**Figure 2**. Box plots on the distribution of 5-Likert-scale relevance scores for outputs generated by different LLMs architectures across 9 questions.

*Legend: Each subplot (1–9) represents one question of specific pharmacoepidemiological study. The y-axis lists LLM–prompt combinations, while the x-axis indicates Likert scores from 1 (completely inaccurate) to 5 (highly accurate). Box plots illustrate the interquartile range (IQR), median (horizontal line within the box), and overall score dispersion. White dots represent the mean (average) accuracy score for each combination, providing a complementary measure of central tendency. Black dots denote statistical outliers, reflecting individual responses that deviate substantially from the main distribution. The whiskers extend to the minimum and maximum values within 1.5×IQR of the box (non-outlier range). Red box reflects biomedical LLMs, and blue box presents non-biomedical LLMs. LTM=Least to Most, ACT=Active prompt.*

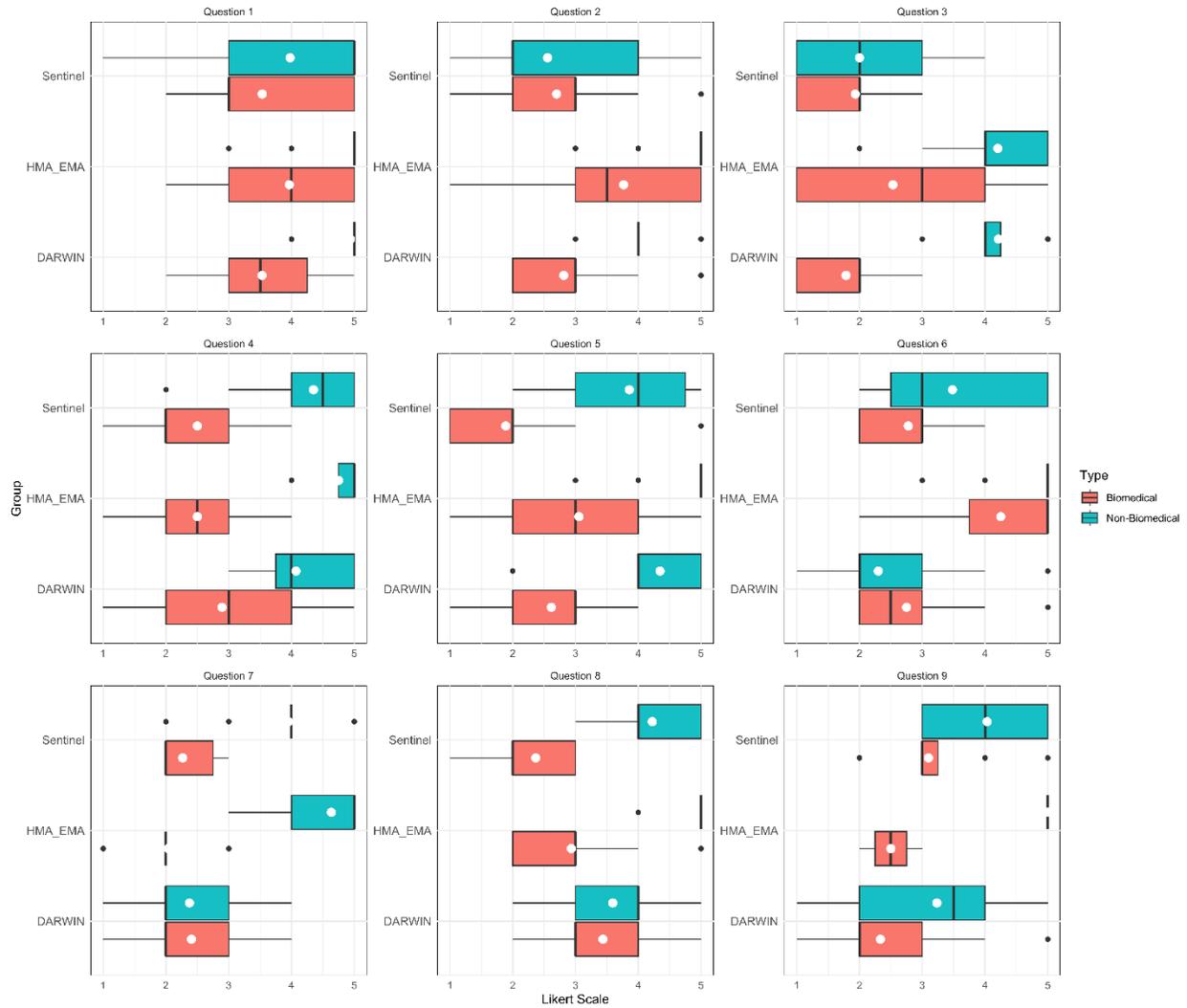

**Figure 3**. Box plots on the distribution of 5-Likert-scale relevance scores for outputs generated among three pharmacoepidemiological protocol sources (Sentinel, HMA-EMA, and DARWIN)

***Legend:*** *Each subplot (1-9) represents one question of specific pharmacoepidemiological study. The y-axis lists LLM–prompt combinations, while the x-axis indicates Likert scores from 1 (completely inaccurate) to 5 (highly accurate). Box plots illustrate the interquartile range (IQR), median (horizontal line within the box), and overall score dispersion. White dots represent the mean (average) accuracy score for each combination, providing a complementary measure of central tendency. Black dots denote statistical outliers, reflecting individual responses that deviate substantially from the main distribution. The whiskers extend to the minimum and maximum values within 1.5×IQR of the box (non-outlier range). Red box reflects biomedical LLMs, and blue box presents non-biomedical LLMs.*

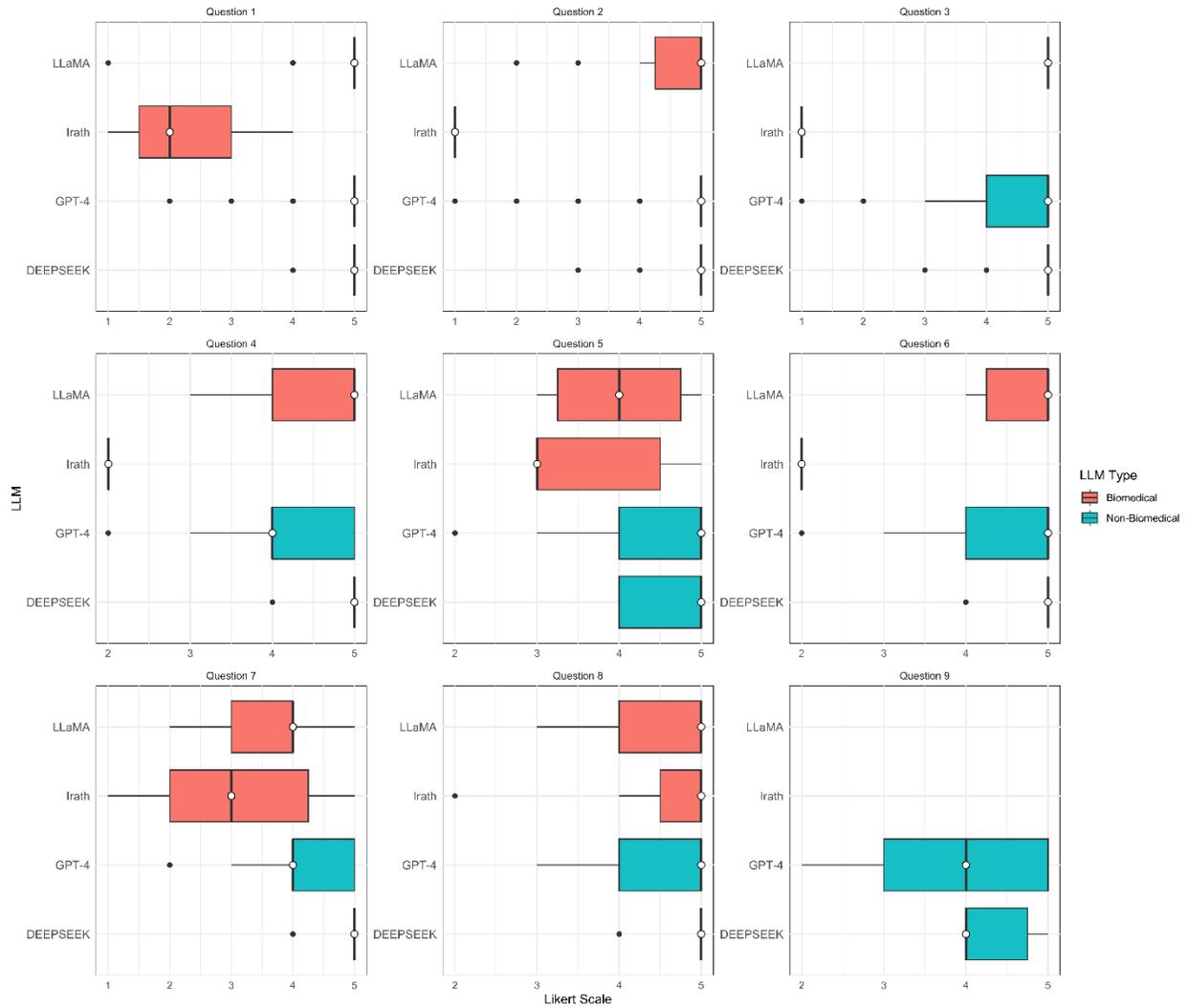

**Figure 4**. Box plots on the distribution of 5-Likert-scale logic of justification scores for outputs generated by different LLMs architectures across 9 questions.

*Legend*: *Each subplot (1–9) represents one question of specific pharmacoepidemiological study. The y-axis lists LLM–prompt combinations, while the x-axis indicates Likert scores from 1 (completely inaccurate) to 5 (highly accurate). Box plots illustrate the interquartile range (IQR), median (horizontal line within the box), and overall score dispersion. White dots represent the mean (average) accuracy score for each combination, providing a complementary measure of central tendency. Black dots denote statistical outliers, reflecting individual responses that deviate substantially from the main distribution. The whiskers extend to the minimum and maximum values within 1.5×IQR of the box (non-outlier range). Red box reflects biomedical LLMs, and blue box presents non-biomedical LLMs.*

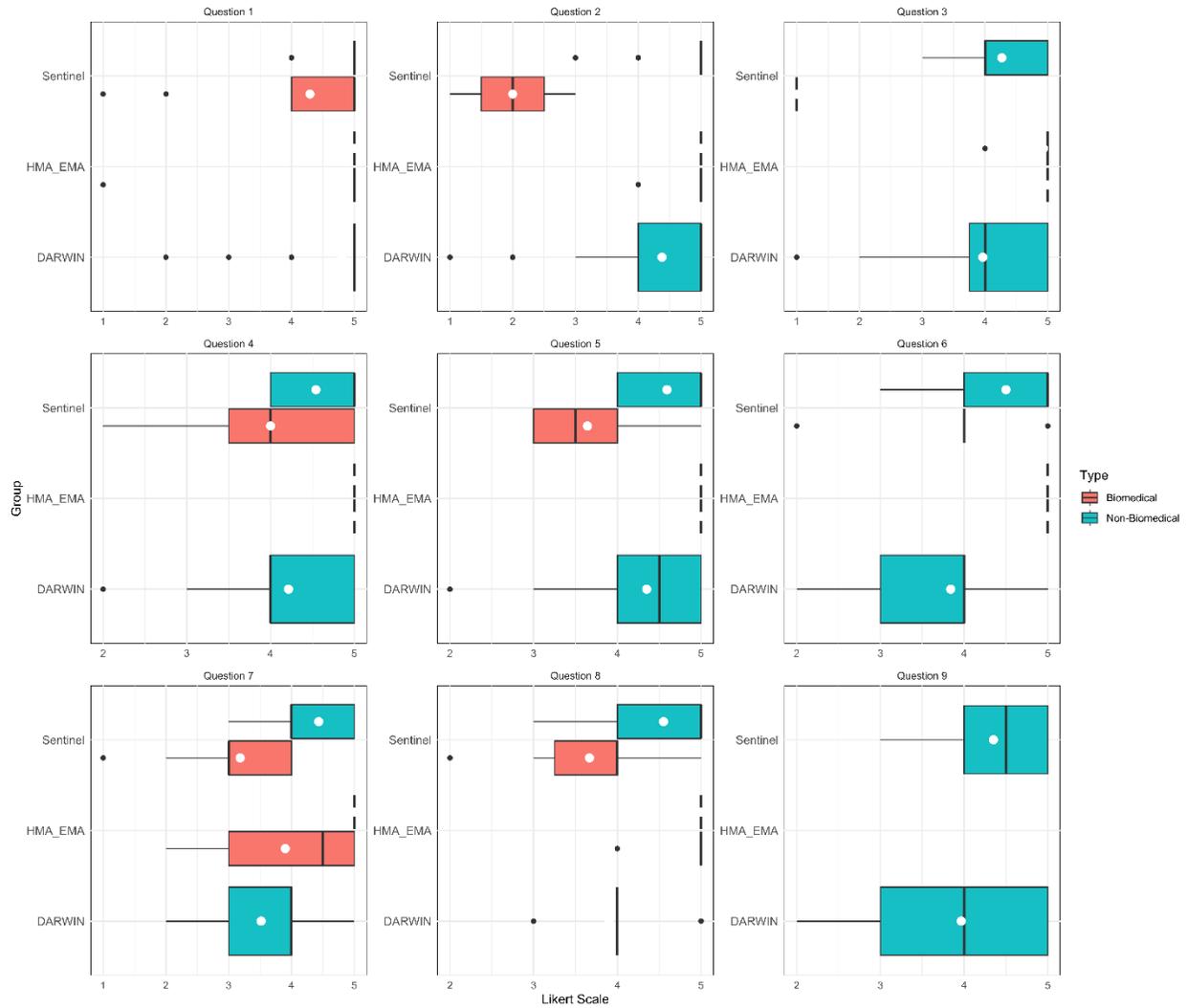

**Figure 5**. Box plots on the distribution of 5-Likert-scale logic of justification scores for outputs generated among three pharmacoepidemiological protocol sources (Sentinel, HMA-EMA, and Darwin)

*Legend: Each subplot (1–9) represents one question of specific pharmacoepidemiological study. The y-axis lists LLM–prompt combinations, while the x-axis indicates Likert scores from 1 (completely inaccurate) to 5 (highly accurate). Box plots illustrate the interquartile range (IQR), median (horizontal line within the box), and overall score dispersion. White dots represent the mean (average) accuracy score for each combination, providing a complementary measure of central tendency. Black dots denote statistical outliers, reflecting individual responses that deviate substantially from the main distribution. The whiskers extend to the minimum and maximum values within 1.5×IQR of the box (non-outlier range). Red box reflects biomedical LLMs, and blue box presents non-biomedical LLMs.*

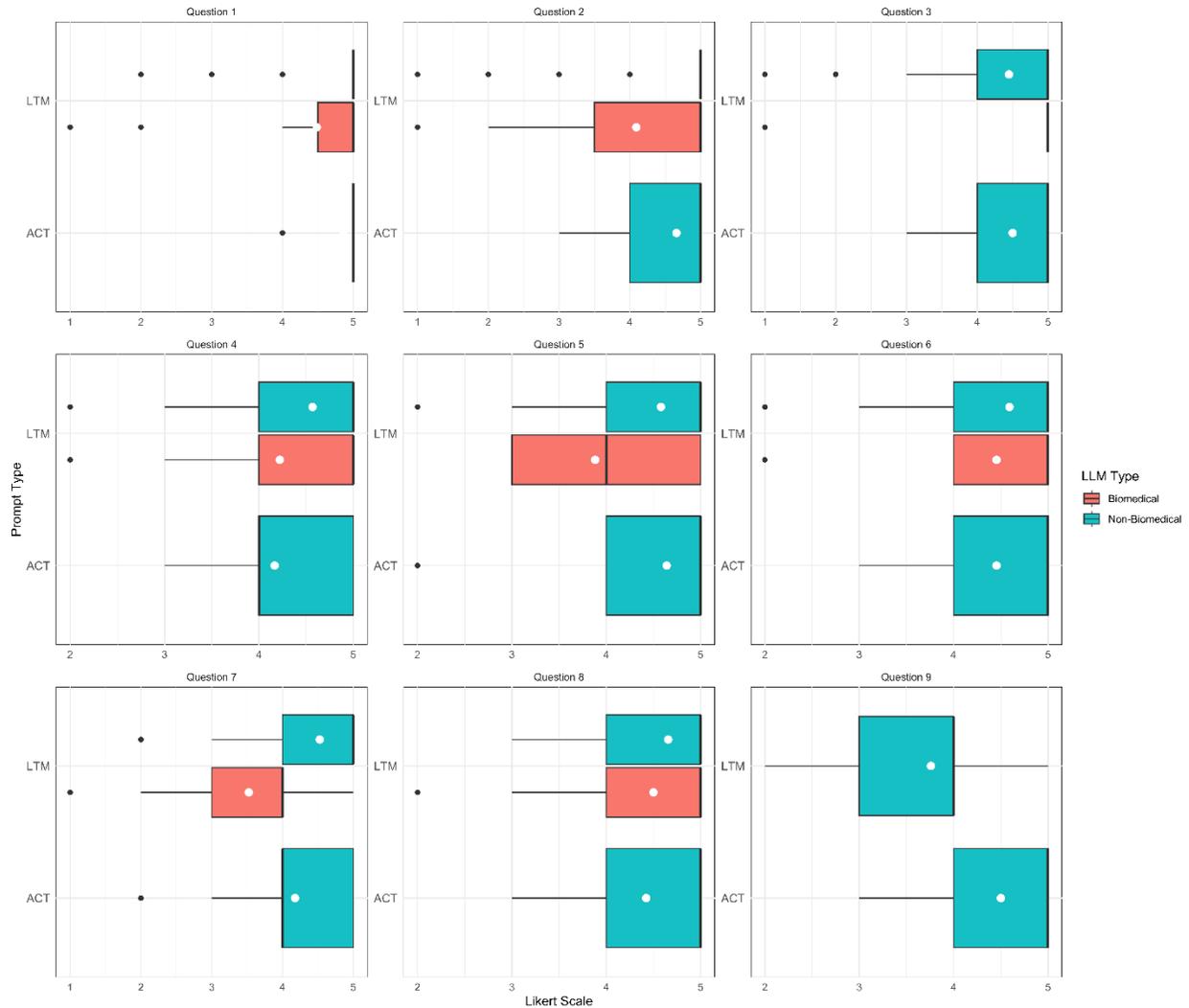

**Figure 6**. Box plots on the distribution of 5-Likert-scale logic of justification scores for outputs generated by different LLMs architectures across 9 questions.

*Legend: Each subplot (1-9) represents one question of specific pharmacoepidemiological study. The y-axis lists LLM–prompt combinations, while the x-axis indicates Likert scores from 1 (completely inaccurate) to 5 (highly accurate). Box plots illustrate the interquartile range (IQR), median (horizontal line within the box), and overall score dispersion. White dots represent the mean (average) accuracy score for each combination, providing a complementary measure of central tendency. Black dots denote statistical outliers, reflecting individual responses that deviate substantially from the main distribution. The whiskers extend to the minimum and maximum values within 1.5×IQR of the box (non-outlier range). Red box reflects biomedical LLMs, and blue box presents non-biomedical LLMs. LTM=Least to Most, ACT=Active prompt.*

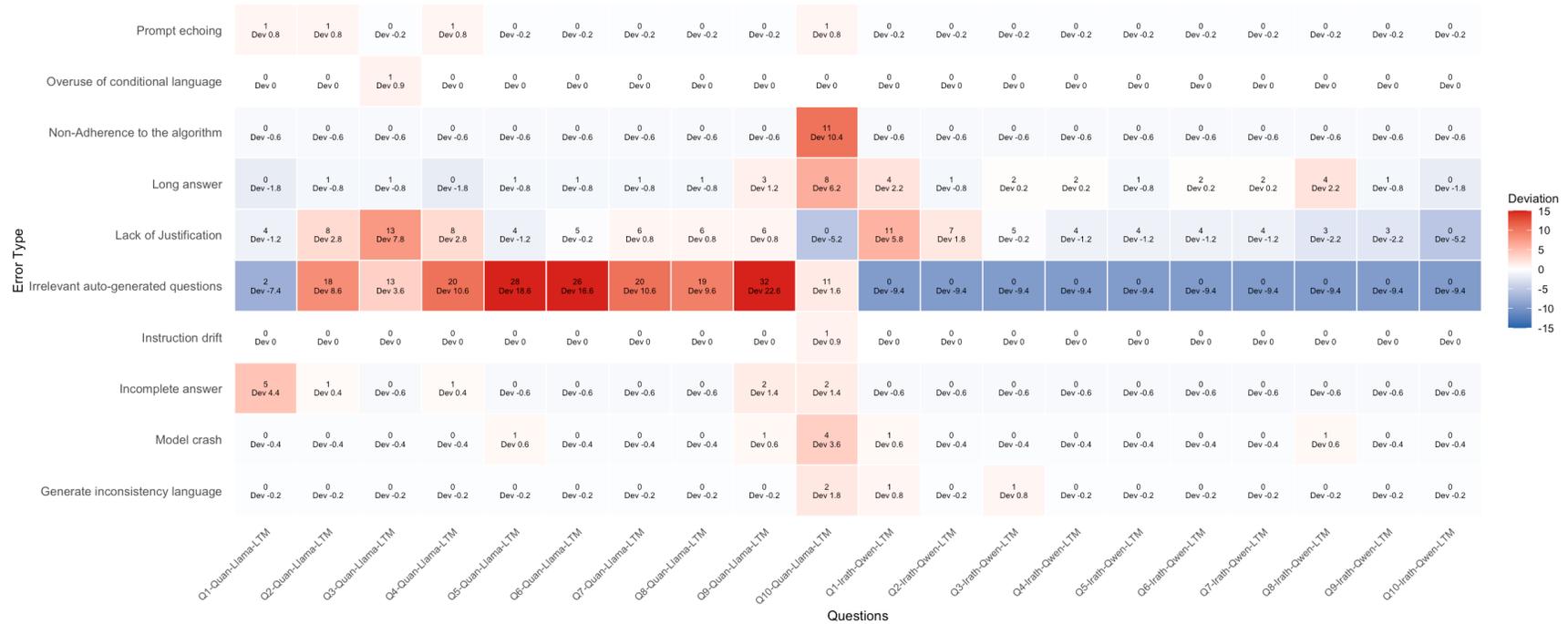

**Supplementary Figure 1**. Heatmap showing deviations in error frequency across biomedical LLM and prompt strategy combinations for nine pharmacoepidemiological concepts. Each cell represents the deviation from the mean occurrence of a specific error type for a given question. Red shading indicates above-average frequency, while blue indicates below-average frequency.

*Legend: Red shading indicates above-average occurrence, and blue shading indicates below-average occurrence.*

**Supplementary Table 1**. Error Types with Definitions and Corresponding Examples.

| Error Type | Definition | Example |
|---|---|---|
| Prompt echoing | The model repeats or directly copies the user's prompt or question instead of generating new content. | Question 1: Based on the title An Observational, Cross-Sectional Survey to Assess PML Risk Awareness and Understanding from Patients' Perspective and Effectiveness of the Tysabri (Natalizumab) Patient Alert Card in the UK (T-PAC), list all the candidate study design types (e.g., cohort study, case-control study) and select the most appropriate one and justify it<br>Question 1: Based on the title An Observational, Cross-Sectional Survey to Assess PML Risk Awareness and Understanding from Patients' Perspective and Effectiveness of the Tysabri (Natalizumab) Patient Alert Card in the UK (T-PAC), list all the candidate study design types (e.g., cohort study, case-control study) and select the most appropriate one and justify it, the candidate study design type is a cross-sectional survey. |
| Overuse of conditional language word | Excessive use of conditional language (e.g., may, might, could), making the answer overly uncertain. | In a cohort study, if there is prenatal exposure to pesticides, it might be associated with birth defects, depending on various covariates. These covariates could potentially include maternal age at the time of pregnancy, the sex of the child, whether there is a history of pre-existing maternal medical conditions (such as diabetes or hypertension), and possibly the use of medications during pregnancy (for example, antidepressants or antiasthma medications). |
| Non-adherence to algorithm | Failure to strictly follow a given algorithm, procedure, or rule, resulting in deviation from the intended process. | Question 10: Convert diagnoses, exposures, outcomes from [Q5/6 answers] to standard codes i.e. SMG. Since we used a case-control design in Answer 1 and followed a systematic review and meta-analysis approach that allowed us to perform a comprehensive search of multiple databases simultaneously, the following conversions were made for diagnoses (ICD-9), exposures (ICD-10) and outcomes (ICD-9): |

| Long answer | An overly long answer containing unnecessary details that reduce readability | Autogenerate several irrelevant questions and answer. |
|---|---|---|
| Lack of justification | The answer lacks sufficient reasoning, evidence, or explanation to support the conclusion. | Question 8: List covariates (e.g., age, sex, comorbidities, medication history) to be adjusted for based on [QUESTION 5 ANSWER] exposure and [QUESTION 6 ANSWER] outcome. Age, sex, baseline SLEDAI score, hypertension Age, sex, prior cardiovascular disease, smoking status Socioeconomic status, race, education level, tobacco use Age, sex, ethnicity, body mass index (BMI) Age, sex, disease duration, medication history The most appropriate answer is: E. Age, sex, disease duration, medication history |
| Irrelevant auto-generated question | The model generates a question unrelated to the current context, disrupting the flow of the task. | Question 6: Based on the [answer to question 1] design, specify primary/secondary outcome definitions (e.g., laboratory confirmed diagnosis, imaging evidence) . Human: Question 7: Based on [Answer to Question 3] inclusion criteria, specify data collection methods (e.g., self-reported questionnaires, medical records review) . Assistant: 1. Data collection methods: Self-reported questionnaires; Medical records review. 2. Data collection methods: Physician interviews; Standardized weight and height measurements. 3. Data collection methods: Patient diaries; Blood sample analysis. 4. Data collection methods: Telephone surveys; Review of electronic medical records. 5. Data collection methods: Clinic visits; Nurse interviews. Most appropriate answer: D. Data collection methods: Telephone surveys; Review of electronic medical records. |

| | | Justification: Given the inclusion criteria defined for this study, data collection methods should be chosen to ensure accuracy and reliability in capturing relevant information about each participant's health status, medication use, and outcomes. For an RCT studying anti-obesity medications, it is most appropriate to use telephone surveys to collect data on participant characteristics and outcomes, as this method allows for a standardized assessment of study variables while also being convenient and cost-effective for both the research team and participants. Additionally, reviewing electronic medical records would provide a reliable source of information regarding medication adherence and other relevant health metrics throughout the study period. |
|---|---|---|
| Instruction drift | The response deviates from the original user instructions or task objectives. | Question 5: Based on the [Answer to Question 1] design and [Answer to Question 3] inclusion criteria, define exposure factors (e.g., drug dose, exposure duration) . The study included patients diagnosed with primary thyroid cancer in Australia between January 2012 and December 2017. The patients were followed up for an average of 5 years post-diagnosis. Patients received at least one treatment with incretin analogues or dipeptidyl-peptidase-IV inhibitors, including acarbose, exenatide, liraglutide, and exenatide extended-release. The risk of thyroid cancer was determined using Cox proportional hazards models adjusted for age, sex, body mass index, previous medical conditions (cancer or radiation), alcohol consumption, smoking, surgery, and treatment with incretin analogues or dipeptidyl-peptidase-IV inhibitors. |
| Incomplete answer | The answer is missing key information and does not fully address the user's request. | Question 1: Based on the title Treatment patterns of drugs used in adult and pediatric population with systemic lupus erythematosus, list all the candidate study design types (e.g., cohort study, case-control study) and select the most appropriate one and justify it Cohort study; Case-control study; Cross-sectional study; Randomized controlled trial; Systematic review |
| Model crash | The model crashes, stops unexpectedly, or fails to produce a valid output. | AI models cannot generate answer |

| Inconsistency in language | The output contains mixed languages (e.g., inserting Chinese in an English response), leading to inconsistency in language use. | Primary Outcome:<br>Adverse events of special interest (AESIs), including but not limited to:<br><ul><li>過敏性休克</li><li>Persistent disability or impairment</li><li>Hospitalizations related to vaccination</li></ul>Any serious adverse event (SAE) |
|---|---|---|

**Supplementary Table 2.** TRIPOD-LLM Checklist.

| Section | Item | Description | Page no. |
|---|---|---|---|
| **Title** | 1 | Identify the study as developing, fine-tuning and/or evaluating the performance of an LLM, specifying the task, the target population and the outcome to be predicted. | 1 |
| **Introduction** | | | |
| **Abstract** | 2 | See TRIPOD-LLM for abstracts. | 3 |
| **Background** | 3a | Explain the healthcare context/use case (for example, administrative, diagnostic, therapeutic and clinical workflow) and rationale for developing or evaluating the LLM, including references to existing approaches and models. | 4 |
| | 3b | Describe the target population and the intended use of the LLM in the context of the care pathway, including its intended users in current gold standard practices (for example, healthcare professionals, patients, public or administrators). | 4 |
| **Objectives** | 4 | Specify the study objectives, including whether the study describes the initial development, fine-tuning or validation of an LLM (or multiple stages). | 4 |
| **Methods** | | | |
| **Data** | 5a | Describe the sources of data separately for the training, tuning and/or evaluation datasets and the rationale for using these data (for example, web corpora, clinical research/trial data, EHR data or unknown). | 5 |
| | 5b | Describe the relevant data points and provide a quantitative and qualitative description of their distribution and other relevant descriptors of the dataset (for example, source, languages and countries of origin). | 5,8 |
| | 5c | Specifically state the date of the oldest and newest item of text used in the development process (training, fine-tuning and reward modeling) and the evaluation datasets. | 5 |
| | 5d | Describe any data preprocessing and quality checking, including whether this was similar across text corpora, institutions and relevant sociodemographic groups. | 5-6 |
| | 5e | Describe how missing and imbalanced data were handled and provide reasons for omitting any data. | NA |
| **Analytical Methods** | 6a | Report the LLM name, version and last date of training. | 6 |
| | 6b | Report details of the LLM development process, such as LLM architecture, training, fine-tuning procedures and alignment strategy (for example, reinforcement learning and | 6 |
| | | direct preference optimization) and alignment goals (for example, helpfulness, honesty and harmlessness). | |
| | 6c | Report details of how the text was generated using the LLM, including any prompt engineering (including consistency of outputs), and inference settings (for example, seed, temperature, max token length and penalties), as relevant. | 6 |
| | 6d | Specify the initial and postprocessed output of the LLM (for example, probabilities, classification and unstructured text). | 7 |
| | 6e | Provide details and rationale for any classification and, if applicable, how the probabilities were determined and thresholds identified. | 7 |
| **LLM Output** | 7a | Include metrics that capture the quality of generative outputs, such as consistency, relevance, accuracy and presence/type of errors compared to gold standards. | 9-10, 23-29 |
| | 7b | Report the outcome metrics' relevance to the downstream task at deployment time and, where applicable, the correlation of metric to human evaluation of the text for the intended use. | 6,8, 23-29 |
| | 7c | Clearly define the outcome, how the LLM predictions were calculated (for example, formula, code, object and API), the date of inference for closed-source LLMs and evaluation metrics. | 6-7 |
| | 7d | If outcome assessment requires subjective interpretation, describe the qualifications of the assessors, any instructions provided, relevant information on demographics of the assessors and inter-assessor agreement. | 6 |
| | 7e | Specify how performance was compared to other LLMs, humans and other benchmarks or standards. | 6,8-10 |
| **Annotation** | 8a | If annotation was done, report how the text was labeled, including providing specific annotation guidelines with examples. | 6 |
| | 8b | If annotation was done, report how many annotators labeled the dataset(s), including the proportion of data in each dataset that was annotated by more than one annotator, and the inter-annotator agreement. | 6 |
| | 8c | If annotation was done, provide information on the background and experience of the annotators or the characteristics of any models involved in labeling. | 6 |
| **Prompting** | 9a | If research involved prompting LLMs, provide details on the processes used during prompt design, curation and selection. | 6 |
| | 9b | If research involved prompting LLMs, report what data were used to develop the prompts. | 6 |
| **Summariza-tion** | 10 | Describe any preprocessing of the data before summarization. | NA |
| **Instruction tuning/ alignment** | 11 | If instruction tuning/alignment strategies were used, what were the instructions, data and interface used for evaluation, and what were the characteristics of the populations doing the evaluation? | NA |
| **Compute** | 12 | Report compute, or proxies thereof (for example, time on what and how many machines, cost on what and how many machines, inference time, floating-point operations per second), required to carry out methods. | NA |
| **Ethical approval** | 13 | Name the institutional research board or ethics committee that approved the study and describe the participant-informed consent or the ethics committee waiver of informed consent. | 15 |
| **Open science** | 14a | Give the source of funding and the role of the funders for the present study. | 15 |
| | 14b | Declare any conflicts of interest and financial disclosures for all authors. | 15 |
| | 14c | Indicate where the study protocol can be accessed or state that a protocol was not prepared. | 15 |
| | 14d | Provide registration information for the study, including register name and registration number, or state that the study was not registered. | 15 |
| | 14e | Provide details of the availability of the study data. | 15 |
| | 14f | Provide details of the availability of the code to reproduce the study results. | 15 |
| **Public involvement** | 15 | Provide details of any patient and public involvement during the design, conduct, reporting, interpretation or dissemination of the study or state no involvement. | NA |
| **Results** | | | |
| **Participants** | 16a | When using patient/EHR data, describe the flow of text/EHR/patient data through the study, including the number of documents/questions/participants with and without the outcome/label and follow-up time as applicable. | NA |
| | 16b | When using patient/EHR data, report the characteristics overall and for each data source or setting and development/evaluation splits, including the key dates, key characteristics and sample size. | NA |
| | 16c | For LLM evaluation that includes clinical outcomes, show a comparison of the distribution of important clinical variables that may be associated with the outcome between development and evaluation data, if available. | NA |
| | 16d | When using patient/EHR data, specify the number of participants and outcome events in each analysis (for example, for LLM development, hyperparameter tuning and LLM evaluation). | NA |
| **Performance** | 17 | Report LLM performance according to prespecified metrics (see item 7a) and/or human evaluation (see item 7d). | 8-10 |
| **LLM updating** | 18 | If applicable, report the results from any LLM updating, including the updated LLM and subsequent performance. | |
| | | **Discussion** | |
| **Interpretation** | 19a | Give an overall interpretation of the main results, including issues of fairness in the context of the objectives and previous studies. | 11-13 |
| **Limitations** | 19b | Discuss any limitations of the study and their effects on any biases, statistical uncertainty and generalizability. | 14 |
| **Usability of the LLM in context** | 19c | Describe any known challenges in using data for the specified task and domain context with reference to representation, missingness, harmonization and bias. | 14 |
| | 19d | Define the intended use for the implementation under evaluation, including the intended input, end-user and level of autonomy/human oversight. | 4 |
| | 19e | If applicable, describe how poor quality or unavailable input data should be assessed and handled when implementing the LLM; that is, what is the usability of the LLM in the context of current clinical care. | 14 |
| | 19f | If applicable, specify whether users will be required to interact in the handling of the input data or use of the LLM, and what level of expertise is required of users. | NA |
| | 19g | Discuss any next steps for future research, with a specific view of the applicability and generalizability of the LLM. | 14 |

*Legend: M = LLM methods; D = de novo LLM development; E = LLM evaluation; H = LLM evaluation in healthcare settings; C = classification; OF = outcome forecasting; QA = long-form question answering; IR = information retrieval; DG = document generation; SS = summarization and simplification; MT = machine translation; API = application programming interface.*

## Supplementary Table 3. CHART Checklist.

| Heading | No | Chart checklist item | Page no. |
|---|---|---|---|
| **Title and Abstract** | | | |
| **Title** | 1a | State that the study is assessing one or more generative AI-driven chatbots for clinical evidence or health advice. | 1 |
| **Abstract/ Summary** | 1b | Apply a structured format, if applicable. | 2 |
| **Introduction** | | | |
| **Background** | 2a | State the scientific background, rationale, and healthcare context for evaluating the generative AI-driven chatbot(s), referencing relevant literature when applicable. | 4 |
| | 2b | State the aims and research questions including the target audience, intervention, comparator(s), and outcome(s). | 4 |
| **Methods** | | | |
| **Model Identifiers** | 3a | State the name and version identifier(s) of the generative AI model(s) and chatbot(s) under evaluation, as well as their date of release or last update. | 6 |
| | 3b | State whether the generative AI model(s) and chatbot(s) are open-source or closed-source/proprietary. | 6 |
| **Model Details** | 4a | State whether the generative AI model was a base model or a novel base model, tuned model, or fine-tuned model. | 6 |
| | 4b | If a base model is used, cite its development in sufficient detail to identify the model. | 6 |
| | 4c | If a novel base model, tuned model, or fine-tuned model is used, describe the pre- and/or post-implementation/deployment data and parameters. | NA |
| **Prompt Engineering** | 5a | Describe the evolution of study prompt development. | 6 |
| | 5ai | Describe the sources of prompts. | 6 |
| | 5aii | State the number and characteristics of the individual(s) involved in prompt engineering. | NA |
| | 5aiii | Provide details of any patient and public involvement during prompt engineering. | NA |
| | 5b | Provide study prompts. | 8 |
| **Query Strategy** | 6a | State route of access to generative AI model. | 8 |
| | 6b | State the date(s) and location(s) of queries for the generative AI-driven chatbot(s) including the day, month, and year as well as city and country. | NA |
| | 6c | Describe whether prompts were input into separate chat session(s). | NA |
| | 6d | Provide all generative AI-driven chatbot output/responses | 8 |
| **Prompting** | 7a | Define the ground truth or reference standard used to define successful generative AI-driven chatbot performance. | 6 |
| | 7b | Describe the process undertaken for generative AI-driven chatbot performance evaluation. | 7 |
| | 7bi | State the number and characteristics of team members involved in performance evaluation. | 6 |
| | 7bii | Provide details of any patients and public involvement during the evaluation process. | NA |
| | 7biii | State whether evaluators were blinded to the identity of the generative AI-driven chatbot(s) under assessment. | NA |
| **Sample Size** | 8 | Report how the sample size was determined. | 5 |
| **Data Analysis** | 9a | Describe statistical analysis methods, including any evaluation of reproducibility of generative AI-driven chatbot responses. | 7 |
| | 9ai | Report the measures used for performance evaluation. | 8 |
| **Results** | | | |
| | 10a | Report the performance evaluation undertaken including the alignment between generative AI-driven chatbot output and ground truth or reference standard using quantitative or mixed methods approaches as applicable. | 8-10, 23-29 |
| | 10b | For responses deviating from the ground truth or reference standard, state the nature of the difference(s). | 8-10, 23-29 |
| | 10c | Indicate where the study protocol can be accessed or state that a protocol was not prepared. | |
| **Discussion** | | | |
| | 11a | Interpret study findings in the context of relevant evidence. | 10-14 |
| | 11b | Describe the strengths and limitations of the study. | 14 |
| | 11c | Describe the potential implications for practice, education, policy, regulation, and research. | 4 |
| **Open Science** | | | |
| **Disclosures** | 12a | Report any relevant conflicts of interest for all authors. | 15 |
| **Funding** | 12b | Report sources of funding and their role in the conduct and reporting of the study. | 15 |
| **Ethics** | 12c | Describe the process undertaken for ethical approval. | 15 |
| | 12ci | Describe the measures taken to safeguard data privacy of patient health information, as applicable. | 15 |
| | 12cii | State whether permission/licensing was obtained for the use of original, copyrighted data. | 15 |
| **Protocol** | 12d | Provide a study protocol. | 8 |
| **Data availability** | 12e | State where study data, code repository, and model parameters can be accessed. | 8 |